%% file: acl_latex.tex
\documentclass[11pt]{article}

\usepackage[preprint]{acl}

\usepackage{times}
\usepackage{latexsym}

\usepackage[T1]{fontenc}
\usepackage[utf8]{inputenc}

\usepackage{microtype}

\usepackage{inconsolata}

\usepackage{graphicx}
\usepackage{xcolor}
\usepackage{listings}
\usepackage{booktabs}
\usepackage{multirow}
\usepackage{multicol}
\usepackage{tabularx}
\usepackage{comment}
\usepackage{tcolorbox}
\tcbuselibrary{listings,skins}

\newcommand{\ignore}[1]{}
\newcommand{\avik}[1]{\textcolor{blue}{Avik: #1}}

\newcommand{\toolcall}{\includegraphics[height=1.2ex]{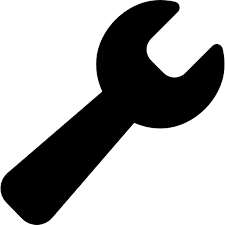}}

\title{An Empirical Investigation of Robustness in Large Language Models under Tabular Distortions}
\author{
Avik Dutta \quad
Harshit Nigam \quad
Hosein Hasanbeig \\
\textbf{Arjun Radhakrishna} \quad
\textbf{Sumit Gulwani} \\
Microsoft Corp. \\
\{avikdutta, harshitnigam, hosein.hasanbeig, arradha, sumitg\}@microsoft.com
}

\begin{document}
\maketitle
\begin{abstract}
Distortions are a natural consequence of tables collected from real-world sources, where scraping, extraction, or manual intervention can introduce semantic and structural inconsistencies. Humans can often identify and mentally correct such errors autonomously before reasoning over the data. In this work, we investigate whether Large Language Models (LLMs) possess a similar capability by evaluating them on 506 distorted table question answering (TQA) samples derived from canonical tables. Our findings reveal systematic weaknesses in current LLM pre-training paradigms, where models exhibit a strong inductive bias toward assuming tables are well-formed, often overlooking distortions or discarding inconsistent entries instead of repairing them.
Even frontier models like GPT-5.4 and Opus-4.6 achieve only 68\% accuracy under distortion.
While distortion-aware prompting partially mitigates these failures, errors persist. 
Through lightweight finetuning, we demonstrate that models can learn to autonomously detect and repair human-recoverable distorted tables.

% We investigate how large language models (LLMs) fail when tabular data in an otherwise canonical representation is subjected to semantic and structural distortions. Our findings reveal that LLMs lack an inherent ability to detect and correct subtle distortions in table representations. Only when provided with an explicit prior, via a system prompt, do models partially adjust their reasoning strategies and correct some distortions, though not consistently or completely.
% To study this phenomenon, we introduce a small, expert-curated dataset that explicitly evaluates LLMs on table question answering (TQA) tasks requiring an additional error-correction step prior to analysis. Our results reveal systematic differences in how LLMs ingest and interpret tabular information under distortion, with even SoTA models such as GPT-5.2 model exhibiting a drop of minimum 22\% accuracy under distortion. These findings raise important questions for future research, particularly regarding when and how models should autonomously decide to realign tabular inputs, analogous to human behavior, without relying on explicit prompts or tabular data pre-processing.
\end{abstract}

\input{content/introduction}

\input{content/methodology}

\input{content/experimental_setup}

\input{content/evaluation}
\input{content/related_works}

\input{content/conclusion}

% Bibliography entries for the entire Anthology, followed by custom entries
%\bibliography{custom,anthology-overleaf-1,anthology-overleaf-2}

\section*{Limitations}
In this study, we focus on single-step distortions rather than multiple compounding transformations. This controlled setup allows us to isolate and analyze the effect of individual distortion patterns without introducing additional confounding factors. Consequently, our evaluation covers only a subset of the much broader space of distorted tables encountered in real-world settings.
Furthermore, we restrict our analysis to distortions that preserve the original answer to the question. This design choice enables us to specifically measure the impact of distortions on model reasoning and table understanding without conflating it with changes in question difficulty or answer computation complexity.
Extending the analysis to cover a broader space remains an important direction for future work.

% Our study is based on a small, expert-curated dataset that focuses on single-step distortions applied to small tables with relatively simple queries. This controlled setting was chosen to isolate distortion-related failures from general task complexity. However, real-world tables are often larger, noisier, and involve multiple interacting errors, which may not fit entirely within the experimental setup we designed.
% Moreover, we restrict our analysis to distortions that preserve cell content in order to keep the original answers valid. While this simplifies evaluation, it excludes content-level corruptions that commonly occur in practice. Extending the analysis to include such distortions is an important direction for future work.

\section*{Ethical Considerations}
Our dataset was constructed by taking samples from WikiTQ \cite{pasupat2015compositionalsemanticparsingsemistructured} which is publicly available under the license CC-BY-SA-4.0\footnote{\url{https://creativecommons.org/licenses/by-sa/4.0/}}.
Human evaluation was conducted across 5 participants from an anonymous IT company.
The experts were compensated on a per-query basis at a rate of \$2 per query.
None of the distortions introduced in this study were designed to target or reinforce biases related to race, gender, ethnicity, or other protected attributes.
This was further approved by the ethics review board by the anonymous IT company. 
All evaluation experiments can be run on a 64GB RAM or a single NVIDIA Tesla V100-32G GPU.
Finetuning was done on a single NVIDIA Tesla T4 (16 GB) GPU.

A key societal concern highlighted by our findings is the risk of \emph{silent failures}, where models generate confident but incorrect outputs under table distortions (Figure~\ref{fig:motivating_eg}). Such failures are particularly dangerous in high-stakes domains like healthcare and finance, where row or column misalignments can alter the meaning of critical values. Our results show that even frontier models remain vulnerable to structural perturbations, emphasizing the need for verification mechanisms and human oversight in safety-critical applications.

ChatGPT was used solely for grammar polishing and improving writing clarity during the preparation of this paper. It was not involved in the formulation of hypotheses, experimental design, analysis, or the scientific discoveries and conclusions presented in this work.

% Custom bibliography entries only
\bibliography{custom}

\appendix
\clearpage
\label{sec:appendix}
\input{content/appendix}

\end{document}

%% file: content/introduction.tex
\section{Introduction}

\begin{figure}[h]
  \includegraphics[width=\columnwidth]{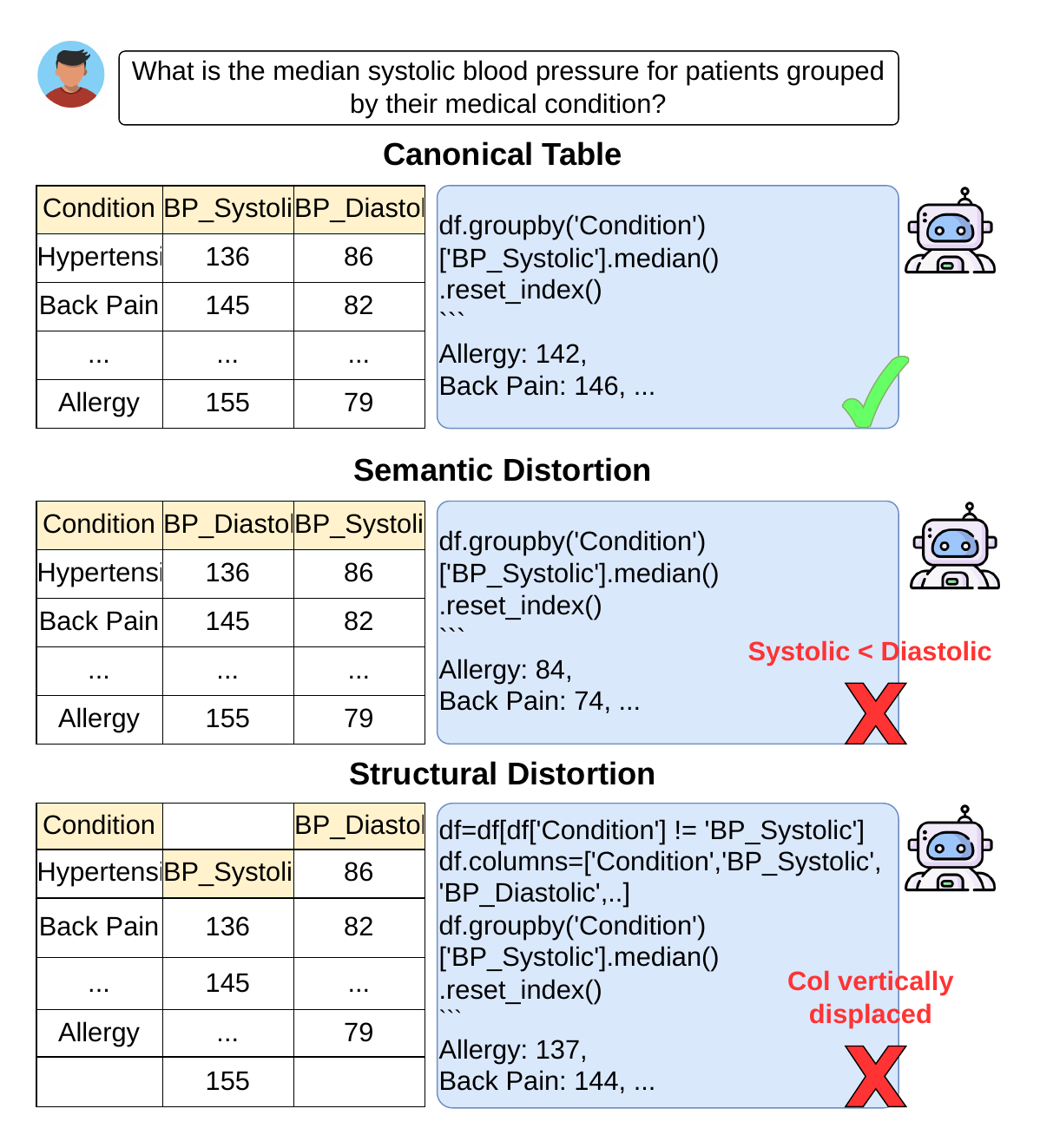}
  \caption{LLMs often overlook distortions in tables. Illustration of \textit{structural} and \textit{semantic} table distortions and their influence on how LLMs reason.}
  \label{fig:motivating_eg}
\end{figure}

LLMs have shown strong performance on TQA tasks, effectively handling both canonical and diverse table representations across different input modalities \cite{zhou2025tablequestionansweringera}.
% \cite{10.1007/978-981-19-7596-7_14}
In practice, however, tables are often imperfect \cite{zhu2025tableevalrealworldbenchmarkcomplex}. Formatting errors, misaligned rows or columns and semantic inconsistencies often arise during data collection, conversion between formats (PDF, Excel) \cite{5277546}, scraping from web \cite{balakrishnan2015applying}, OCR from images, etc. While such tables may appear “broken,” they often retain a latent internal structure that humans can easily recognize and repair before using them for downstream applications (Figure~\ref{fig:motivating_eg}).
In this work, we investigate if LLMs can autonomously detect and correct such "weird but solvable" tables. 

Our study shows error-aware reasoning over tables is not a learned default behavior: when tables are distorted, model accuracy degrades sharply, even when the distortions are visually or semantically evident. Only when models are provided with an explicit prior, via distortion-aware system prompts, do they adjust their reasoning strategies, and even then, recovery is inconsistent and incomplete.
To systematically study this phenomenon, we curate a dataset of 506 TQA tasks designed to isolate distortion-related failures. Each canonical table is transformed into multiple \textit{semantic} and \textit{structural} distorted variants, while preserving the original answer, allowing us to directly measure robustness under controlled errors. Through evaluations across model families, input modalities, and execution settings, we uncover consistent failure patterns, most notably a widespread inability to handle vertical structural shifts that require reasoning about global table layout.
Our findings suggest a fundamental gap in current table understanding capabilities: \textbf{LLMs either fail to detect or treat distortions as noisy inputs to be cleaned rather than as misaligned structures to be repaired.} 
To alleviate this inductive bias in LLMs, we demonstrate supervised finetuning with distortion-aware reasoning traces can encourage models to inspect table structure, question computation validity, and autonomously repair inconsistencies --- behaviors that humans perform inherently.

% This raises important questions about how future models should decide when to pre-process, realign, or question tabular inputs, behaviors that humans routinely perform without explicit instruction.

\ignore{
\avik{Problems that currently need addressing to improve the paper.
\begin{enumerate}
    \item run statistical significance tests on the results.
    \item Go over the rebuttal and see if there is anything that still needs to be added.
\end{enumerate}
}

\avik{Our goal is to investigate and suggest startegies instead of showing ful-proof solutions on how this can be solved.}
}

\ignore{

We judge if LLMs are robust to distorted layouts or "weird but solvable" tables by reversing the error logic. These are some representation styles that do not appear in typical training corpora.

We intentionally include broken tables that have latent internal structure which a human can fix with reasoning. However, when the same problem is projected to an LLM, can they invert the error to reconstruct the true form to provide the right solution for the given query.

From our experiments, we observe that LLMs often do not fully comprehend the underlying meaning of the data and its context, and instead tend to directly apply standard solution patterns to arrive at an answer.

\avik{Release the dataset under an anonymouse github repo.}
}

%% file: content/methodology.tex
\section{Distortions}
We define distortions as operational transformations applied to canonical representations that introduce errors or inconsistencies that makes tables logically incorrect.
Unlike noise, which is random and unstructured and therefore difficult to recover from, distortions are patterned perturbations that preserve the underlying semantic information and can therefore be repaired or recovered.
% Distortion differs from noise in that distortions are structured or patterned perturbations that often preserve underlying semantic information and can therefore be corrected or recovered from. In contrast, noise is typically random, unstructured, and semantically uninformative, making it difficult to recover the original signal and instead requiring removal or filtering.
% Unlike standard preprocessing operations that normalize values under a fixed schema to remove noise, distortions disrupt the schema itself, either structurally or semantically, requiring explicit inference and repair before downstream reasoning can proceed. 
For the scope of this paper, we restrict ourselves to single operational distortions under two headings -- \textbf{semantic} and \textbf{structural}.

\subsection{Semantic Distortion}
Semantic Distortions alter the semantic interpretation of a table by introducing logical or relational inconsistencies.
For example, the second table in Figure~\ref{fig:motivating_eg} contains a \textit{BP\_Systolic} value lower than \textit{BP\_Diastolic}, which violates real-world medical constraints.
Consequently, beyond solving the task, LLMs must also identify such inconsistencies using world knowledge and domain understanding.
% Typical semantic distortions considered in this study include: (i) character misrepresentation (e.g., $5 \leftrightarrow S$); (ii) insertion of stray characters (e.g., \texttt{2`35}); (iii) decimal separator swaps (e.g., \texttt{2,345.65} $\leftrightarrow$ \texttt{2.345,65}); and (iv) label mismatches (e.g., weight measured in cubic centimeters).
Typical semantic distortions considered in this study include: (i) \textit{CharMis}: character misrepresentation (e.g., $5 \leftrightarrow S$); (ii) \textit{StrChar}: insertion of stray characters (e.g., \texttt{2`35}); (iii) \textit{DecSwap}: decimal separator swaps (e.g., \texttt{2,345.65} $\leftrightarrow$ \texttt{2.345,65}); and (iv) \textit{LabMis}: label mismatches (e.g., weight measured in cubic centimeters).
More in Appendix~\ref{appx:semantic_distortion_type}.
% In this type of distortion, we purposely introduce a mistake which can affect the semantic meaning of how the table is read and interpreted. 
% This allows us to assess whether LLMs move beyond surface-level assumptions and instead verify table content using world knowledge and relational constraints.
% Some of the principles we followed during distortions are by breaking known logical or numerical invariants. 
% For instance in Figure \ref{fig:motivating_eg} we marked the higher blood pressure column as \textit{BP\_Diastolic} and lower blood pressure column as \textit{BP\_Systolic} which inverts the semantic meaning of the two columns.
% For example, in Figure \ref{fig:motivating_eg}, \textit{BP\_Diastolic} and \textit{BP\_Systolic} are swapped, inverting their semantic meaning.
% Other principles we followed include label value mismatch (\textit{Age} containing 120/80), inconsistent scale or unit (Weight measured in cubic centimeters, Volume in kilograms), etc.

\subsection{Structural Distortion}
Structural distortion disrupts the spatial alignment between headers, rows, and cells, thereby introducing inconsistencies in the table structure. For instance, in the third table of Figure~\ref{fig:motivating_eg}, the \textit{BP\_Systolic} column is vertically shifted downward, causing misalignment with the remaining columns. To correctly solve such tables, the evaluated model must reason about the underlying table geometry and verify whether the structural alignment is semantically consistent before performing computation.
Typical structural distortions considered in this study include:
(i) \textit{Vert}: vertical shift, where a subset of columns is vertically misaligned relative to the rest; (ii) \textit{Horz}: horizontal shift, where certain rows are horizontally displaced relative to the headers; (iii) \textit{CMerge}: content merge, where multiple row, column or cell values are incorrectly merged into a single entry; (iv) \textit{CSplit}: content split, where a single entry is fragmented across multiple rows or columns; and (v) \textit{Foot}: footnote injection, where footnote markers are inserted within the table content.
More details in Appendix~\ref{appx:structural_distortion_type}.

% In this type of distortion, we maintain the semantics but alter the spatial alignment between headers, rows and cells to create disruption. 

% Some of the principles we followed are: vertical shifts of columns (Figure~\ref{fig:motivating_eg}), Horizontal row displacements (\textit{Condition} missing, \textit{BP\_Systolic} appearing under \textit{BP\_Diastolic}), single cells split (does the model merge them or treat them independently), headers embedded inside data, etc.

\begin{table*}[t]
\footnotesize
\centering
\begin{tabularx}{\textwidth}{ll*{9}{r}}
    \toprule
    \multirow{3}*{\textbf{Model}} & \multirow{3}*{\textbf{Input}} & \multirow{3}*{\textbf{Can.}} & \multicolumn{4}{c}{\textbf{Dist. Unaware}} & \multicolumn{4}{c}{\textbf{Dist. Aware}} \\
    \cmidrule(l){4-7}\cmidrule(l){8-11}
    & & & \multicolumn{3}{c}{\textbf{Distorted}} & \textbf{Rbst.} & \multicolumn{3}{c}{\textbf{Distorted}} & \textbf{Rbst.} \\
    \cmidrule(l){4-6} \cmidrule(l){8-10}
    & & & Sem$^\dagger$ & Str$^\ddagger$ & \textbf{All} &  & Sem$^\dagger$ & Str$^\ddagger$ & \textbf{All} &  \\
    \midrule
    
    \multicolumn{11}{c}{\textbf{Table-Finetuned Large Language Models}} \\
    \midrule
    TableLLM-7B  & Text \toolcall & 34.79 & 12.00 & 3.97 & 7.76 & 22.30 & 11.40 & 3.53 & 7.25 & 20.84 \\
    TableGPT2-7B & Text \toolcall & 46.96 & 14.41 & 7.63 & 10.84 & 23.08 & 19.07 & 10.73 & 14.69 & 31.28 \\

    \midrule
    \multicolumn{11}{c}{\textbf{Open-sourced Large Language Models}} \\
    \midrule
    Qwen2.5-VL-32B & Text \toolcall  & 53.66 & 24.90 & 19.32 & 21.98 & 40.96 & 27.92 & 17.87 & 22.66 & 42.23 \\
    Qwen2.5-VL-32B & Image \toolcall & 51.11 & 24.17 & 15.97 & 19.88 & 38.90 & 20.83 & 18.32 & 19.52 & 38.19 \\

    Deepseek-R1-Distill {\tiny Qwen-32B} & Text & 74.70 & 60.42 & 50.57 & 55.25 & 73.96 & 60.83 & 51.52 & 55.95 & 74.90 \\

    \midrule

    \multicolumn{11}{c}{\textbf{Close-sourced Large Language Models}} \\
    \midrule

    GPT-5.4 & None \toolcall  & 83.96 & 68.05 & 61.45 & 64.61 & 76.95 & 82.57 & 67.05 & 74.46 & 88.69 \\
    GPT-5.4 & Text            & 81.42 & 67.63 & 54.72 & 60.87 & 74.76 & 72.20 & 56.98 & 64.23 & 78.89 \\
    GPT-5.4 & Text \toolcall  & 83.99 & 69.71 & 60.75 & 65.02 & 77.41 & 82.16 & 66.79 & 74.11 & 88.24 \\
    GPT-5.4 & Image           & 70.36 & 63.90 & 53.21 & 58.30 & 82.86 & 67.50 & 53.58 & 60.20 & 85.56 \\
    GPT-5.4 & Image \toolcall & 81.62 & 69.58 & 62.26 & 65.74 & 80.54 & 78.01 & 66.04 & 71.74 & 87.89 \\

    Claude-Opus-4.6 & None \toolcall  & 88.32 & 80.91 & 69.81 & 75.10 & 85.03 & 87.97 & 78.87 & 83.20 & 94.20 \\
    Claude-Opus-4.6 & Text            & 87.98 & 73.22 & 71.76 & 72.46 & 82.36 & 78.33 & 74.72 & 76.44 & 86.88 \\
    Claude-Opus-4.6 & Text \toolcall  & 87.55 & \textbf{84.23} & \textbf{73.96} & \textbf{78.85} & \textbf{90.06} & 90.46 & \textbf{78.93} & \textbf{84.46} & \textbf{96.47} \\
    Claude-Opus-4.6 & Image           & 85.94 & 66.67 & 59.92 & 63.13 & 73.46 & 73.84 & 62.88 & 68.06 & 79.19 \\
    Claude-Opus-4.6 & Image \toolcall & 87.35 & 83.40 & 68.30 & 75.49 & 86.42 & \textbf{91.25} & 77.36 & 83.96 & 96.12 \\
    
    \bottomrule
\end{tabularx}
\caption{
\textbf{Depicts accuracy degradation (\%) when answering identical queries over canonical versus distorted tables.} Distortions are categorized into semantic (Sem$^\dagger$) and structural (Str$^\ddagger$). Robustness ($\textit{Rbst}\%=\textit{Dist.}/\textit{Can.}$) quantifies performance preservation under distortion. Models are evaluated across \textit{None}, \textit{Text} and \textit{Image} modes, with and without code sandbox (\toolcall). Variance across 5 iterations $\approx 0.6857$.
% Results indicate that current LLMs lack inherent robustness to table distortions, though distortion-aware prompting yields consistent improvements. 
}
\label{tab:results}
\end{table*}

\section{Dataset Construction}
We construct our dataset\footnote{Full dataset available at \url{https://github.com/AIML-Researcher/table-distortion}} by programmatically introducing controlled distortions that change the structural layout and semantic consistency of tables.
We generate a collection of 506 \textit{(table, query, answer)} distortions derived from 316 canonical tables from WikiTQ \cite{pasupat2015compositionalsemanticparsingsemistructured}.
The distorted samples are subsequently validated through a round of LLM-as-Judge followed by Human evaluation, yielding 265 structurally distorted and 241 semantically distorted instances (Appendix~\ref{appx:dataset_construction}).
Importantly, the original query--answer pairs are preserved throughout the transformation process, ensuring that our evaluation isolates the ability of LLMs to identify, reason about, and recover from distortions, rather than being confounded by variations in question complexity or answer distribution.

%% file: content/experimental_setup.tex
\section{Experimental Setup}

\paragraph{Models} Our evaluation covers LLMs over three broad categories: 
(1) \textbf{Finetuned}: TableGpt2-7B \cite{su2024tablegpt2largemultimodalmodel}, TableLLM-7B \cite{zhang2025tablellmenablingtabulardata};
(2) \textbf{Open-source}: Qwen2.5-VL-32B \cite{qwen2.5-VL}, Deepseek-R1-Distill-Qwen-32B \cite{deepseekai2025deepseekr1incentivizingreasoningcapability};
(3) \textbf{Closed-source}: GPT-5.4 (reasoning: medium) \cite{openai2026gpt54}, Claude-Opus-4.6 \cite{anthropic2026opus46}\footnote{Models accessed on 05/24/26 under default settings}.

\paragraph{Setup.} 
The above models are evaluated both with and without access to a Docker-based code execution sandbox that supports file uploads and Python execution. This setup enables a fair comparison between models that rely primarily on direct reasoning and those that solve tasks through code execution. For models without native tool- or function-calling capabilities, generated Python scripts are executed separately within the sandbox, and the output used for evaluation.
Furthermore, tabular context is provided under three different modalities--(1) \textbf{None}: the table file is directly uploaded to the code sandbox without giving any preview in the prompt; (2) \textbf{Text}: a markdown representation is provided; (3) \textbf{Image}: a PNG image of the entire table is provided.
This helps us study whether input modality influences how models detect and handle distortions.
We use Markdown instead of other textual forms, like CSV, as it better preserves the structural orientations we cover in the dataset.

% \avik{We decided not to include TableLlava because of its low performance even on the canonical representation. Mistral also hallucniates on a non-sandboxed environment. deepseek tooling is omitted since the python programs.}

\paragraph{Metrics.} 
We report performance using exact-match \cite{pasupat2015compositionalsemanticparsingsemistructured}. Additionally, we measure Robustness (\%) as the ratio of accuracy on distorted tables to accuracy on their corresponding canonical representations. This metric captures the extent to which model performance is preserved under distortion, while also enabling fair comparison across models with varying baseline accuracies.

%% file: content/evaluation.tex
\section{Evaluation}

Table~\ref{tab:results} reports model accuracy for both Canonical and Distorted forms. 
Table-Finetuned LLMs exhibit the weakest robustness, showing negligible recovery under structural distortions, suggesting that table headers and row-level samples as context during training do not help detect layout-level information.
Among open-source models, Qwen2.5 shows moderate robustness ($\approx40\%$), while DeepSeek performs substantially better (>50\%). Closed-source models such as GPT-5.4 and Opus-4.6 perform strongest overall, but still reach only up to 78.85\% accuracy under distortion.
Across input modalities, Text generally achieves higher accuracy than Image, with this gap being more prominent for structural than semantic, suggesting that current image-based pretraining approaches are ineffective in capturing structural irregularities in tables.

\subsection{Prompt Steering}
We incorporate distortion-handling instructions (exact prompt in Appendix~\ref{appx:prompt_details}) in the system prompt and evaluate LLMs.
Table-Finetuned and Open-sourced models show little to no improvement, indicating limited capability despite explicit awareness.
In contrast, closed-source models demonstrate substantially stronger distortion-awareness, improving average robustness by 7.35\% for GPT-5.4 and 7.11\% for Opus-4.6 across all input modalities, reaching a best of 96.47\% by the latter.
However, the gains are considerably larger for semantic than for structural (+7.64\% versus +4.71\%), suggesting that structural inconsistencies remain significantly more challenging to detect and repair.
This trend is particularly evident in vertical column shifts, where even the best-performing configuration, Opus-4.6 (Text \toolcall), achieves only 57.38\% (Table~\ref{tab:distortion_breakdown_mistake}).

\subsection{Distortion-aware Finetuning}

Motivated by the limitations of prompt steering on Open-sourced models, we fine-tuned Qwen2.5-1.5B-Instruct and Gemma-3-4B-it using supervised fine-tuning (SFT) with QLoRA \cite{dettmers2023qloraefficientfinetuningquantized} for 3 epochs under a 5-fold cross validation setup with 80/20 train--test split. All experiments are conducted on a single NVIDIA Tesla T4 (16 GB) GPU.
Our training set mixes canonical and distorted tables so the model learns to detect distortions before repairing them, rather than assuming every table is corrupted. Training outputs follow a structured format: the model inspects the table for structural and semantic consistency, repairs any inconsistencies it finds while computing, and then gives the answer (see Appendix~\ref{appx:finetuning_exp}).

\begin{table}[h]
\footnotesize
\centering
\begin{tabular}{lccccc}
\toprule
\textbf{Model} & \textbf{Can.} & \multicolumn{3}{c}{\textbf{Distorted}} & \textbf{Rbst.} \\
\cmidrule(l){3-5}
& & \textbf{Sem$^\dagger$} & \textbf{Str$^\ddagger$} & \textbf{All} & \\
\midrule
\multicolumn{5}{c}{\textbf{Qwen2.5-1.5B-Instruct}} \\
\midrule
Base &
26.09 &
18.97 &
17.65 &
18.48 &
70.83 \\

Finetuned &
\textbf{31.52} &
\textbf{22.07} &
\textbf{25.29} &
\textbf{23.26} &
\textbf{73.79} \\

&
\scriptsize{$\pm$ 1.33} &
\scriptsize{$\pm$ 5.64} &
\scriptsize{$\pm$ 9.44} &
\scriptsize{$\pm$ 6.55} &
-- \\

\midrule
\multicolumn{5}{c}{\textbf{Gemma-3-4B-it}} \\
\midrule
Base &
32.61 &
27.59 &
35.29 &
30.43 &
\textbf{93.31} \\

Finetuned &
\textbf{55.87} &
\textbf{37.24} &
\textbf{42.35} &
\textbf{39.13} &
70.04 \\

&
\scriptsize{$\pm$ 5.01} &
\scriptsize{$\pm$ 5.12} &
\scriptsize{$\pm$ 8.97} &
\scriptsize{$\pm$ 3.99} &
-- \\

\bottomrule
\end{tabular}
\caption{Finetuning improves success under distortion.}
\label{tab:finetuning_main}
\end{table}

Table~\ref{tab:finetuning_main} shows that lightweight finetuning improves distortion handling, with gains of 4.78\% for Qwen2.5 and 8.70\% for Gemma-3 over their base models. Accuracy on canonical tables also improves, suggesting that models learnt robust reasoning rather than memorizing distortion patterns.
Per distortion (Table~\ref{tab:finetuning_distortion_type}), Qwen2.5 recovers on \textit{label mismatch}, \textit{content merge}, and \textit{content split}, which the base model failed entirely, while Gemma-3 gains most on \textit{stray characters insertion} (+45.00) and \textit{horizontal shift} (+30.91). 
Gains are uneven (e.g., drops in \textit{vertical shift} and \textit{content split} for Gemma-3), and the high standard deviations indicate that some folds fail to learn these signals consistently, highlighting the sensitivity of finetuning in low-resource settings. Nevertheless, the overall improvements suggest that models can be trained to detect and repair table distortions autonomously.

\begin{table}[t]
\footnotesize
\centering
\setlength{\tabcolsep}{3pt}
\begin{tabularx}{\columnwidth}{Xrr@{\hspace{12pt}}rr}
\toprule
\textbf{Distortions} &
\multicolumn{2}{c}{\textbf{Qwen2.5-1.5B}} &
\multicolumn{2}{c}{\textbf{Gemma-3-4B}} \\
\cmidrule(lr){2-3}
\cmidrule(lr){4-5}
& \textbf{Base} & \textbf{Finetuned} & \textbf{Base} & \textbf{Finetuned} \\
\midrule

\multicolumn{5}{c}{\textbf{Semantic Distortions}} \\
\midrule
CharMis
& 17.24 & 15.17 {\scriptsize$\pm$ 6.26}
& 20.69 & 32.41 {\scriptsize$\pm$ 9.32} \\

StrChar
& 25.00 & 30.00 {\scriptsize$\pm$ 11.18}
& 50.00 & 95.00 {\scriptsize$\pm$ 11.18} \\

LabMis
& 0.00 & 20.00 {\scriptsize$\pm$ 14.49}
& 22.22 & 20.00 {\scriptsize$\pm$ 9.30} \\

DecSwap
& 31.25 & 33.75 {\scriptsize$\pm$ 5.59}
& 37.50 & 41.25 {\scriptsize$\pm$ 5.59} \\

\midrule
\multicolumn{5}{c}{\textbf{Structural Distortions}} \\
\midrule

Vert
& 16.67 & 26.67 {\scriptsize$\pm$ 14.91}
& 66.67 & 33.33 {\scriptsize$\pm$ 0.00} \\

Horz
& 45.45 & 36.36 {\scriptsize$\pm$ 17.01}
& 36.36 & 67.27 {\scriptsize$\pm$ 13.79} \\

CMerge
& 0.00 & 14.55 {\scriptsize$\pm$ 10.37}
& 18.18 & 30.91 {\scriptsize$\pm$ 15.21} \\

CSplit
& 0.00 & 28.00 {\scriptsize$\pm$ 10.95}
& 40.00 & 32.00 {\scriptsize$\pm$ 30.33} \\

Foot
& 0.00 & 0.00 {\scriptsize$\pm$ 0.00}
& 0.00 & 0.00 {\scriptsize$\pm$ 0.00} \\

\bottomrule
\end{tabularx}
\caption{Comparing base and finetuned Qwen2.5-1.5B and Gemma-3-4B-it models across distortion types.}
\label{tab:finetuning_distortion_type}
\end{table}

%% file: content/related_works.tex
\section{Related Works}

Prior work has extensively studied table question answering \cite{chen-etal-2020-hybridqa, yavuz-etal-2018-takes} focusing on how models reason over tabular content under different representations \cite{sui2024tablemeetsllmlarge, wu2025realhitbenchcomprehensiverealistichierarchical} and input modalities \cite{zheng2024multimodaltableunderstanding, yang2025doestablesourcematter}, typically assuming that tables are canonical and well-formed \cite{Herzig_2020, liu2022tapextablepretraininglearning, jiang2022omnitabpretrainingnaturalsynthetic}.
More recent studies have begun to examine robustness by introducing controlled perturbations \cite{bhandari2025exploringrobustnesslanguagemodels}, while RobuT \cite{zhao-etal-2023-robut} adversarially alter table rows and columns without changing the underlying semantics to evaluate model stability. Other works compare performance across alternative but valid table representations for the same query \cite{zhang2025contentdifferentrepresentationscontrolled}, or introduce explicit mechanisms to filter noise and spurious query clauses \cite{ye2025tableqameetsnoisedual}.
We study rule-based errors and how they can be mitigated towards autonomous rectification than rejection.

%% file: content/conclusion.tex
\section{Conclusion}
Our evaluation reveals that current LLMs exhibit a strong inductive bias toward assuming tables are well-formed, which prevent them from repairing both semantic and structural inconsistencies. While distortion-aware prompting improves robustness, recovery remains incomplete and inconsistent. Through lightweight fine-tuning, we demonstrate that these limitations can be alleviated.

% provides insights into how LLMs are trained to process table-based content. Finetuned and open-source models exhibit a strong inductive bias toward assuming tables are correct, while closed-source models show greater flexibility but still make systematic errors. Across all models, vertical shifts remain a major failure mode, revealing limited understanding of global table structure. These results highlight the need for future table understanding systems that can autonomously detect and handle distorted tables rather than assuming well-formed inputs.

%% file: content/appendix.tex
\begin{figure*}[htbp]
  \includegraphics[width=\textwidth]{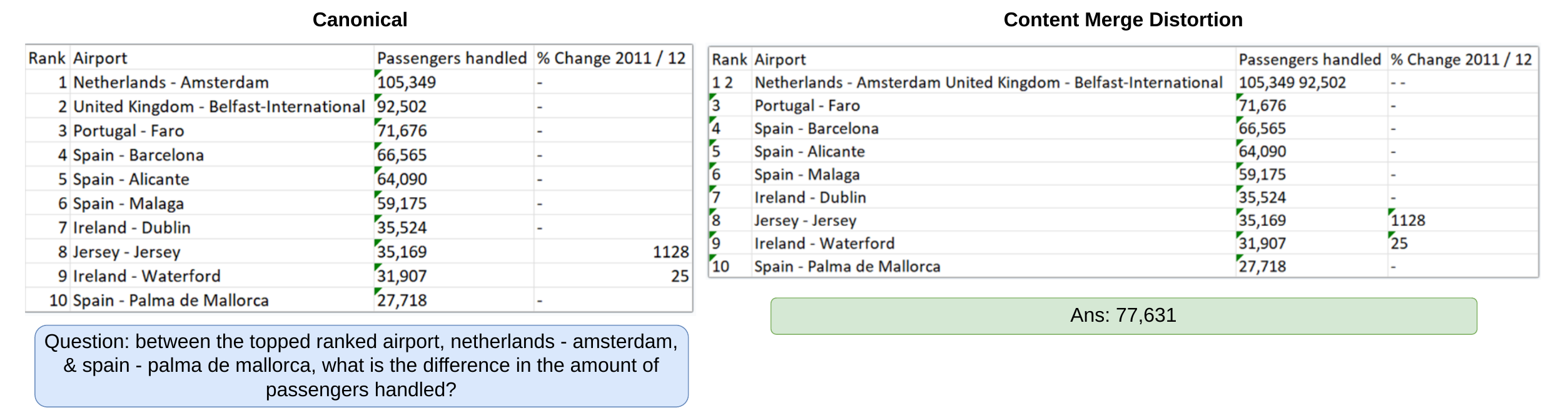}
  \caption{The model must realize that the first and the second row have been merged together and they are not a single row data.}
  \label{fig:content_merge}
\end{figure*}
%%%%%%%%%%%%%%%%%%%%%%%%%%

%%%%%%%%%%%%%%%%%%%%%%%%%%
\begin{figure*}[htbp]
  \includegraphics[width=\textwidth]{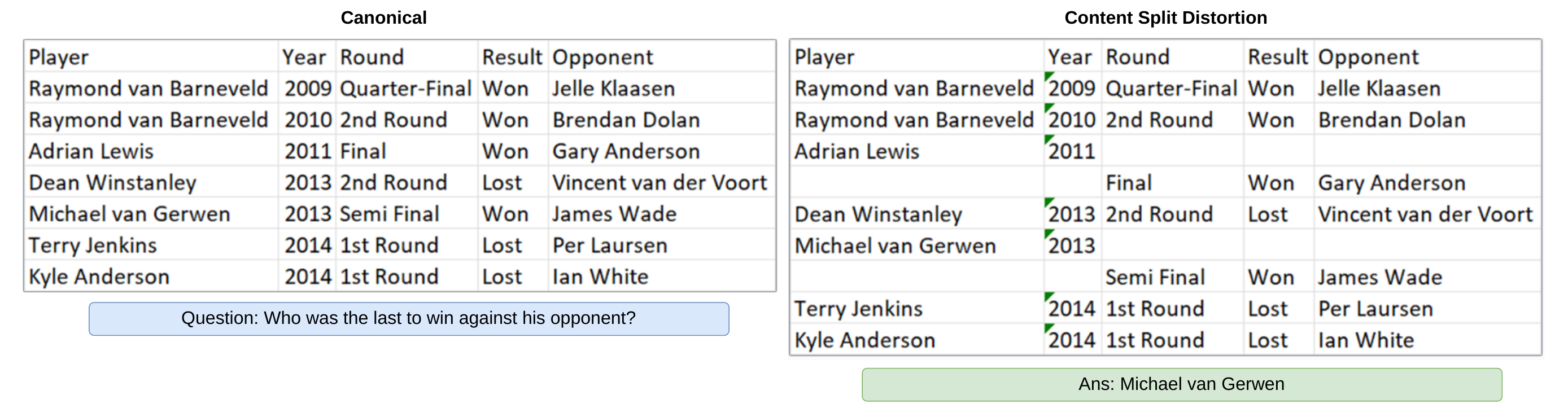}
  \caption{The model should align the displacement of partial columns into different rows correctly. This should not be interpreted as new data with missing fields.}
  \label{fig:content_split}
\end{figure*}
%%%%%%%%%%%%%%%%%%%%%%%%%%

% \begin{figure*}[htbp]
%   \includegraphics[width=\textwidth]{image/footnote_injection.drawio.pdf}
%   \caption{\textbf{Structural Distortion Example 2:} The model must realize that P400 labels under ProductName do not make sense unless the the rows have been horizontally displaced.}
%   \label{fig:footnote_injection}
% \end{figure*}

%%%%%%%%%%%%%%%%%%%%%%%%%%
\begin{figure*}[htbp]
  \includegraphics[width=\textwidth]{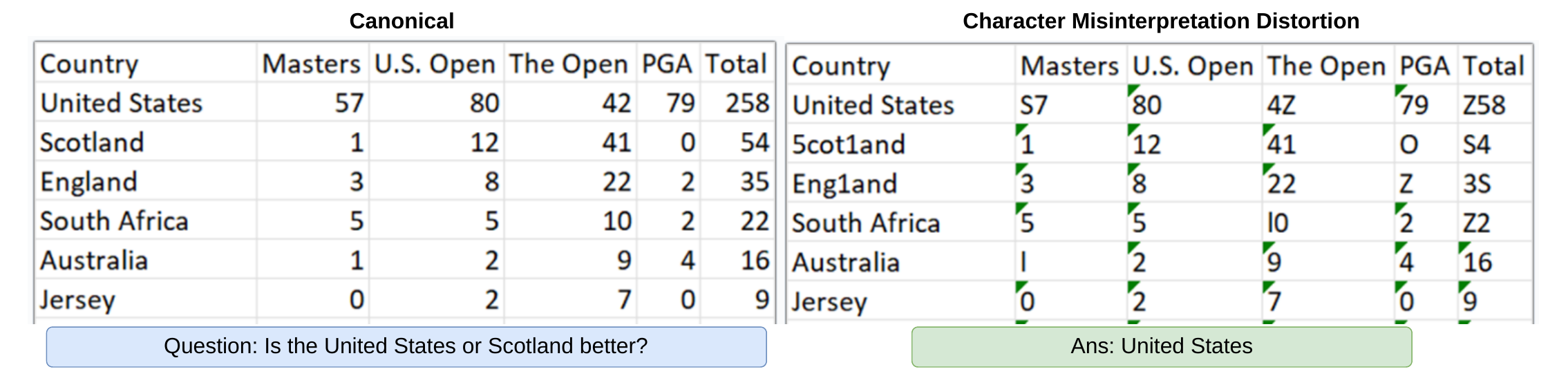}
  \caption{The model must correctly map the country names and numerical values and not simply use a greater than logic and compute directly on the string.}
  \label{fig:character_misinterpretation}
\end{figure*}

\section{Dataset Construction}
\label{appx:dataset_construction}
In this section, we discuss the specifics of how tables from WikiTQ were transformed to create a verified distorted set of 506 samples for evaluation. 
We sampled 500 \textit{(table, query, answer)} triplets from the validation split from WikiTQ and applied controlled programmatic transformations to insert different distortion types at locations that would impact the answer computation targets in the table.
The result, 3891 different deterministic disturbances that are multiple variants of the same canonical table, were passed on to an LLM-as-Judge stage to review and filter transformations that: (i) does not violate the internal semantics or implied meaning of the table, (ii) retains the query--answer relevance from the canonical representation and alters the computation strategy for the given answer.
We used Claude-Opus-4.5 model for this purpose. Stage (i) filtered out 1195 valid samples out of 3891 (69.3\% rejected).
Stage (ii) kept only distortions that materially altered the computation strategy for the answer: 1195 --> 640 (46.4\% rejected). This proceeded to human validation.

\paragraph{Human Annotation.} Five annotators from an anonymous IT company validated the 640 samples using the same guidelines as the LLM judge, yielding 506 accepted samples (20.94\% rejected). For inter-annotator agreement, we asked the annotators to label 100 samples such that each test had at least two annotations. While the annotators reached a raw agreement of 83\%, the average Cohen Kappa score was low (0.11). Because the candidate set had already passed LLM-judge filtering, the label distribution was highly skewed towards those to be retained \cite{humanannotation}. We therefore report Gwet's AC1, a prevalence-robust coefficient, as our primary reliability measure: AC1 = 0.79, indicating substantial agreement and trust on the benchmark set used for evaluation.

\section{Distortion Types}
\label{appx:distortion_types}
In this section, we describe the various distortions applied to canonical tables in detail. These distortions commonly arise in tables curated or extracted from image and text documents, as well as from errors introduced during manual data entry. While the exact transformations vary depending on the table content, they can be broadly categorized into the following groups.

\begin{table}[h]
\centering
\begin{tabular}{lr}
\toprule
\textbf{Distortion Type} & \textbf{Percentage (\%)} \\
\midrule
\multicolumn{2}{c}{\textbf{Semantic}} \\
\midrule
character\_misinterpretation & 19.96 \\
stray\_characters\_insertion & 7.51 \\
label\_mismatch              & 6.32 \\
decimal\_separator\_swap     & 13.83 \\
\midrule
\multicolumn{2}{c}{\textbf{Structural}} \\
\midrule
vertical\_shift              & 12.06 \\
horizontal\_shift            & 11.07 \\
content\_merge               & 18.77 \\
content\_split               & 6.13 \\
footnote\_injection          & 4.35 \\
\bottomrule
\end{tabular}
\caption{Distribution of distortion types across the 506 samples curated for this study. Values represent percentage of the total samples.}
\label{tab:diversification_distribution}
\end{table}

\subsection{Semantic}
\label{appx:semantic_distortion_type}
We cover 4 broad categories where tables in the wild can be distorted semantically.

\paragraph{character\_misinterpretation}
Characters in cell values are swapped with visually similar ones (OCR-style confusions, e.g.\ O$\leftrightarrow$0, l$\leftrightarrow$1, rn$\leftrightarrow$m). See Figure~\ref{fig:character_misinterpretation}.

\paragraph{stray\_characters\_insertion}
Random noise characters or symbols are inserted into otherwise valid table values. For example, artifacts such as stains or marks in scanned documents may appear as symbols like an asterisk ($*$) within numeric entries, which are subsequently captured by OCR systems. This could also confuse an LLM in altering the scale of values, where for a certain label that only accepts whole numbers gets a floating point number in between.

\paragraph{label\_mismatch}
Labels do not align with their corresponding values, or rules which break relation constraints inside the table. For example, systolic pressure scoring lower than diastolic pressure in Figure~\ref{fig:motivating_eg}.

\paragraph{decimal\_separator\_swap}
Decimal and thousands separators are swapped (e.g.\ 1,234.56 $\leftrightarrow$ 1.234,56), changing the numeric reading of a value without touching the digits themselves.

\subsection{Structural}
\label{appx:structural_distortion_type}
We cover 5 broad categories where tables obtained in the wild can be distorted structurally.

\paragraph{vertical\_shift}
Rows are nudged up or down so that values end up under the wrong row label (the row dimension is misaligned with its data).

\paragraph{horizontal\_shift}
Columns are nudged left or right so values fall under the wrong header (the column dimension is misaligned with its data).

\paragraph{content\_merge}
Multiple cells, rows, or columns are fused into one, collapsing information that was originally separated (see example Figure~\ref{fig:content_merge}).

\paragraph{content\_split}
A single logical table entity is broken across multiple physical rows or columns, fragmenting a record that should stay together (see example Figure~\ref{fig:content_split}).

\paragraph{footnote\_injection}
Footnote markers, asterisks, or trailing reference notes are inserted into cells, polluting the value with non-data text. These usually get picked from pdf files or documents when the table spans multiple pages and when page number, header or footer can creep in.

%%%%%%%%%%%%%%%%%%%%%%%%%%%%%%%%%%%%%%%%%%%%%%%%%%%

\section{Evaluation across Distortion Types}

We report robustness across individual distortion categories in Tables~\ref{tab:distortion_breakdown} and~\ref{tab:distortion_breakdown_mistake}. 
The results reveal substantial differences not only across model families, but also between semantic and structural perturbations.

Table-finetuned models continue to perform poorly across nearly all distortion categories. 
In the distortion unaware setting, both TableLLM and TableGPT2 achieve near-zero accuracy for several structural perturbations, particularly horizontal shifts and content splits. 
Even under distortion-aware prompting, improvements remain limited and are largely confined to semantic distortions. 
For example, TableGPT2 improves noticeably on semantic corruptions such as stray character insertion and label mismatch, but structural robustness remains extremely weak, with content split accuracy remaining fixed at only 3.23\% and vertical shift accuracy stagnating at 6.67\%. 
This further suggests that prompt adjustments do not help close the fundamental gaps of how these models were finetuned. 

Among open-source models, Qwen2.5-VL and DeepSeekR1 exhibit moderate robustness gains under distortion-aware prompting, though their weaknesses remain distortion-specific. 
Vertical shifts remain particularly challenging across the board: Qwen2.5-VL achieves only 11.67\% accuracy even in the aware setting, while DeepSeekR1 remains below 32\%. 
Qwen additionally struggles with structural perturbations more broadly, especially horizontal shifts and content merges, where performance remains below 20\% despite prompting. 
In contrast, DeepSeekR1 handles horizontal shifts substantially better, achieving close to 78\% accuracy, suggesting stronger spatial reasoning capabilities. 
However, prompting yields only marginal gains overall, indicating that awareness alone is insufficient to reliably repair structural inconsistencies.

Closed-source frontier models demonstrate significantly stronger robustness across nearly all distortion types. 
Across GPT-5.4 and Opus-4.6 variants, semantic distortions such as character misinterpretation, label mismatch, and decimal separator swaps are handled reliably, with many settings exceeding 85--90\% accuracy. 
Structural perturbations, however, remain comparatively harder. 
In particular, vertical shifts and content merges consistently emerge as the weakest categories across both model families. 
Even in distortion-aware settings, GPT-5.4 variants average only around 20-40\% robustness on vertical shifts, while Opus-4.6, although substantially stronger, still shows noticeably lower performance on this distortion compared to other categories.

Distortion-aware prompting nevertheless provides meaningful gains for frontier models, particularly on semantic corruptions and localized structural inconsistencies. 
For instance, Opus-4.6 nearly saturates on horizontal shifts and character-level semantic distortions, with several configurations reaching or exceeding 90\% robustness. 
Similarly, GPT-5.4 shows strong improvements on content merges and content splits under aware prompting. 
Interestingly, these gains are not uniform: vertical shift robustness slightly decreases for some GPT-5.4 configurations in the aware setting, suggesting that prompting can occasionally bias the model toward over-correction or erroneous table restructuring. 
Overall, the results indicate that while large frontier models can effectively utilize distortion-aware instructions, structural reasoning, particularly row and column alignment under global table perturbations, remains an unresolved challenge even for the strongest systems evaluated.

Furthermore, the choice of input modality noticeably affects robustness for specific distortion types. Text-based inputs consistently outperform image-based settings on semantic distortions such as label mismatch, stray character insertion, and decimal separator swaps, particularly for open-source models like Qwen2.5-VL, suggesting that OCR-induced noise compounds existing semantic corruption. In contrast, image inputs occasionally help preserve coarse structural layout information, leading to competitive or slightly improved performance on distortions such as horizontal shifts and content splits for some closed-source models. GPT-5.4 and Opus-4.6 generally achieve their strongest overall robustness under text-based prompting, while the ``None'' setting remains surprisingly competitive for semantic distortions, indicating that frontier models can often infer and repair localized inconsistencies without explicit tabular content guidance. However, these gains do not extend uniformly across all structural perturbations. Vertical shifts remain consistently difficult across text, image, and prompt-aware settings alike, highlighting that maintaining global row-column alignment and relational consistency is a fundamental challenge that persists regardless of evaluation framework or modality.

\begin{table*}[!t]
\footnotesize
\centering
\begin{tabularx}{\textwidth}{ll*{9}{r}}
    \toprule
    \multirow{2}*{\textbf{Model}} & \multirow{2}*{\textbf{Input}} 
    & \multicolumn{4}{c}{\textbf{Semantic}} & \multicolumn{5}{c}{\textbf{Structural}} \\
    
    \cmidrule(l){3-6} \cmidrule(l){7-11}

    & & \textbf{CharMis} & \textbf{StrChar} & \textbf{LabMis} & \textbf{DecSwap} 
    & \textbf{Vert} & \textbf{Horz} & \textbf{CMerge} & \textbf{CSplit} & \textbf{Foot} \\
    
    \midrule
    \multicolumn{11}{c}{\textbf{Table-Finetuned Large Language Models}} \\
    \midrule
    TableLLM & Text & 10.53 & 11.43 & 15.62 & 12.70 & 3.39 & 0.00 & 6.32 & 0.00 & 9.52 \\   TableGPT2 & Text & 6.00 & 29.73 & 21.88 & 14.93 & 6.67 & 3.70 & 11.58 & 3.23 & 9.09 \\

    \midrule
    \multicolumn{11}{c}{\textbf{Open-source Large Language Models}} \\
    \midrule
    Qwen2.5VL  & Text \toolcall  & 10.89 & 55.26 & 28.12 & 27.14 & 9.84 & 14.29 & 23.40 & 19.35 & 40.91 \\
    Qwen2.5VL  & Image \toolcall & 13.86 & 47.37 & 31.25 & 23.19 & 3.28 & 12.50 & 21.51 & 22.58 & 27.27 \\
    DeepseekR1 & Text            & 53.00 & 73.68 & 43.75 & 71.43 & 27.87 & 78.57 & 43.16 & 58.06 & 63.64 \\

    \midrule
    \multicolumn{11}{c}{\textbf{Close-sourced Large Language Models}} \\
    \midrule
    GPT-5.4 & None \toolcall  & 61.39 & 81.58 & 71.88 & 68.57 & 40.98 & 76.79 & 58.51 & 89.66 & 54.55 \\
    GPT-5.4 & Text            & 66.34 & 84.21 & 59.38 & 64.29 & 26.23 & 85.71 & 43.16 & 87.10 & 59.09 \\
    GPT-5.4 & Text \toolcall  & 64.36 & 89.47 & 53.12 & 74.29 & 36.07 & 73.21 & 56.84 & 93.55 & 68.18 \\
    GPT-5.4 & Image           & 71.29 & 81.58 & 43.75 & 52.86 & 29.51 & 69.64 & 47.37 & 80.65 & 63.64 \\
    GPT-5.4 & Image \toolcall & 61.39 & 84.21 & 70.97 & 72.86 & 36.07 & 82.14 & 57.89 & 90.32 & 63.64 \\

    Opus-4.6 & None \toolcall  & 70.30 & 92.11 & 81.25 & 90.00 & 57.38 & 82.14 & 65.26 & 87.10 & 68.18 \\
    Opus-4.6 & Text            & 67.00 & 83.78 & 75.00 & 75.71 & 37.70 & 98.18 & 70.53 & 93.33 & 76.19 \\
    Opus-4.6 & Text \toolcall  & 79.21 & 94.74 & 75.00 & 90.00 & 52.46 & 100.00 & 67.37 & 87.10 & 77.27 \\
    Opus-4.6 & Image           & 59.00 & 80.00 & 62.50 & 72.86 & 37.70 & 83.64 & 52.69 & 77.42 & 68.18 \\
    Opus-4.6 & Image \toolcall & 77.23 & 92.11 & 78.12 & 90.00 & 52.46 & 80.36 & 64.21 & 87.10 & 72.73 \\

    \bottomrule
\end{tabularx}
\caption{Breakdown of model robustness across individual distortion categories in a \textbf{Distortion Unaware} setting.}
\label{tab:distortion_breakdown}
\end{table*}

%%%%%%%%%%%%%%%%%%%%%%%%%%%%%%%%%%%%%%%%%%%%%%%%%%%%%%%%%%

\begin{table*}[!t]
\footnotesize
\centering
\begin{tabularx}{\textwidth}{ll*{9}{r}}
    \toprule
    \multirow{2}*{\textbf{Model}} & \multirow{2}*{\textbf{Input}} 
    & \multicolumn{4}{c}{\textbf{Semantic}} & \multicolumn{5}{c}{\textbf{Structural}} \\
    
    \cmidrule(l){3-6} \cmidrule(l){7-11}

    & & \textbf{CharMis} & \textbf{StrChar} & \textbf{LabMis} & \textbf{DecSwap} 
    & \textbf{Vert} & \textbf{Horz} & \textbf{CMerge} & \textbf{CSplit} & \textbf{Foot} \\
    
    \midrule
    \multicolumn{11}{c}{\textbf{Table-Finetuned Large Language Models}} \\
    \midrule

    TableLLM & Text \toolcall & 9.38 & 14.29 & 18.75 & 9.23 & 3.39 & 2.00 & 5.26 & 0.00 & 4.76 \\
    TableGPT2 & Text \toolcall & 8.00 & 37.84 & 31.25 & 19.40 & 6.67 & 3.77 & 18.95 & 3.23 & 13.64 \\
    
    \midrule
    \multicolumn{11}{c}{\textbf{Open-source Large Language Models}} \\
    \midrule
    Qwen2.5VL  & Text \toolcall  & 15.84 & 60.53 & 31.25 & 26.09 & 11.67 & 14.29 & 18.95 & 23.33 & 31.82 \\
    Qwen2.5VL  & Image \toolcall &  7.92 & 55.26 & 28.12 & 17.39 &  6.67 & 16.07 & 17.20 & 45.16 & 22.73 \\
    DeepseekR1 & Text            & 53.47 & 78.95 & 50.00 & 66.67 & 31.15 & 78.18 & 44.21 & 54.84 & 68.18 \\

    \midrule
    \multicolumn{11}{c}{\textbf{Close-sourced Large Language Models}} \\
    \midrule
    GPT-5.4 & None \toolcall  & 80.20 & 94.74 & 71.88 & 84.29 & 42.62 & 83.93 & 67.02 & 90.32 & 59.09 \\
    GPT-5.4 & Text            & 79.21 & 84.21 & 59.38 & 61.43 & 22.95 & 82.14 & 49.47 & 90.32 & 72.73 \\
    GPT-5.4 & Text \toolcall  & 79.21 & 94.74 & 68.75 & 85.71 & 36.07 & 91.07 & 65.26 & 90.32 & 63.64 \\
    GPT-5.4 & Image           & 79.21 & 78.95 & 48.39 & 52.86 & 27.87 & 75.00 & 48.42 & 83.87 & 50.00 \\
    GPT-5.4 & Image \toolcall & 77.23 & 86.84 & 71.88 & 77.14 & 34.43 & 85.71 & 67.37 & 83.87 & 72.73 \\

    Opus-4.6 & None \toolcall  & 85.15 & 94.74 & 81.25 & 91.43 & 70.49 & 92.86 & 73.68 & 90.32 & 72.73 \\
    Opus-4.6 & Text            & 76.24 & 94.59 & 78.12 & 72.86 & 44.26 & 98.21 & 73.68 & 90.32 & 81.82 \\
    Opus-4.6 & Text \toolcall  & 91.09 & 94.74 & 81.25 & 91.43 & 57.38 & 100.00 & 74.73 & 93.55 & 81.82 \\
    Opus-4.6 & Image           & 70.41 & 83.78 & 65.62 & 77.14 & 42.62 & 91.07 & 55.32 & 70.97 & 68.18 \\
    Opus-4.6 & Image \toolcall & 94.00 & 94.74 & 81.25 & 90.00 & 63.93 & 92.86 & 73.68 & 90.32 & 72.73 \\

    \bottomrule
\end{tabularx}
\caption{Breakdown of model robustness across individual distortion categories in a \textbf{Distortion Aware} setting.}
\label{tab:distortion_breakdown_mistake}
\end{table*}

%%%%%%%%%%%%%%%%%%%%%%%%%%%%%%%%%%%%%%%%%%%%%%%%%%

\section{Finetuning}
\label{appx:finetuning_exp}
We describe the implementation details of finetuning on the distorted table samples.

\paragraph{Training Setup.}
We fine-tuned \textit{Qwen2.5-1.5B-Instruct} and \textit{Gemma-3-4B-it} using supervised fine-tuning (SFT) on a mixture of canonical and distorted tables. The 506 distorted samples were first split into an 80/20 train--test partition. Each distorted table was mapped back to its corresponding canonical representation, and both versions were assigned to the same split to avoid leakage. This resulted in a final dataset of 824 samples, with 647 samples used for 5-fold cross-validation during training (518 train and 129 validation samples per fold) and 177 held out for testing. Using both clean and distorted tables prevents the model from overfitting to explicit distortion patterns while encouraging it to distinguish between well-formed and corrupted tables.

Training targets followed a structured reasoning format consisting of three primary stages: \textsc{SCAN}, where the model checks whether the table structure appears logically consistent; \textsc{COMPUTE}, where it derives the answer; and \textsc{CONCLUDE}, where it states the final response. Two additional optional stages were introduced that fire only when the table warrants them: \textsc{FLAG}, if something does not look right, and \textsc{REPAIR}, that applies a fix for the flagged error. 
Distorted samples follow either one of the following trajectories during reasoning: \textit{upfront} (the anomaly is obvious at first glance, so FLAG fires during SCAN), \textit{mid-computation} (the anomaly only becomes apparent when reading the cell the question depends on, so FLAG fires during COMPUTE), or \textit{none} (the distortion does not affect the cells needed to answer the question and is therefore not mentioned).
Canonical samples follow \textit{none} by default.
This allows flexibility for the model to detect and de-distort even if something was not apparent superficially in the \textsc{SCAN} stage.
When building the target output for training, during \textsc{REPAIR}, we provide only the relevant de-distortion definition for the current sample to reduce memorization and overfitting to other types.

\begin{figure}[h]
  \includegraphics[width=\columnwidth]{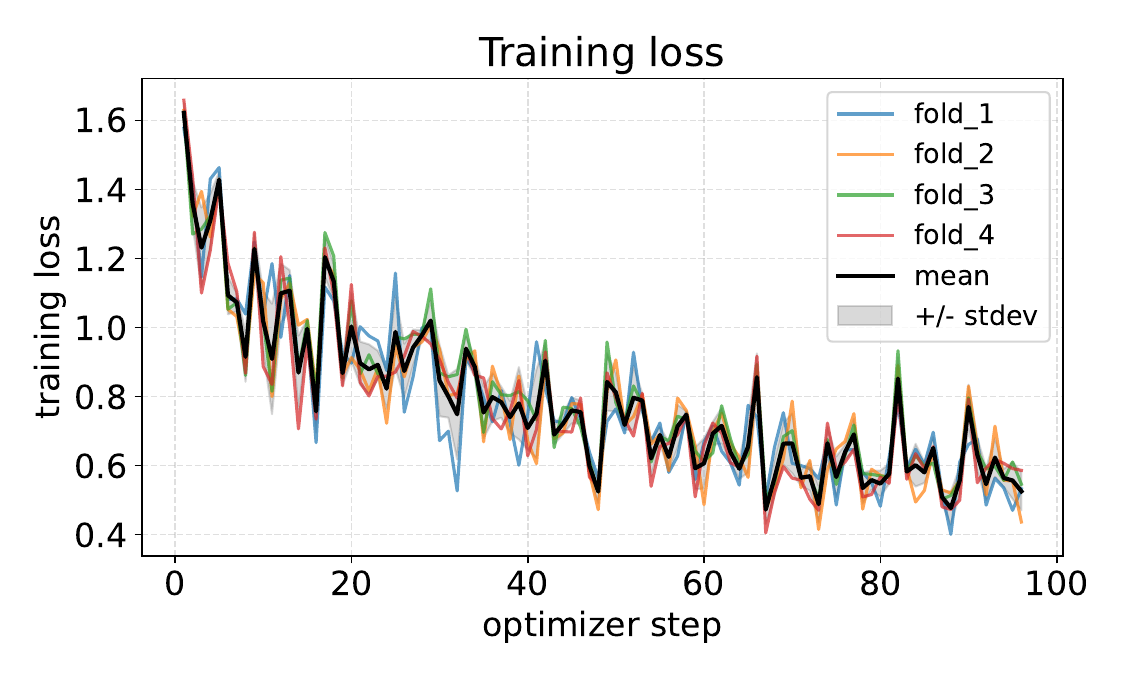}
  \caption{Fine-tuning loss curve on Qwen2.5-1.5B-Instruct.}
  \label{fig:finetuning_loss}
\end{figure}

\begin{figure}[h]
  \includegraphics[width=\columnwidth]{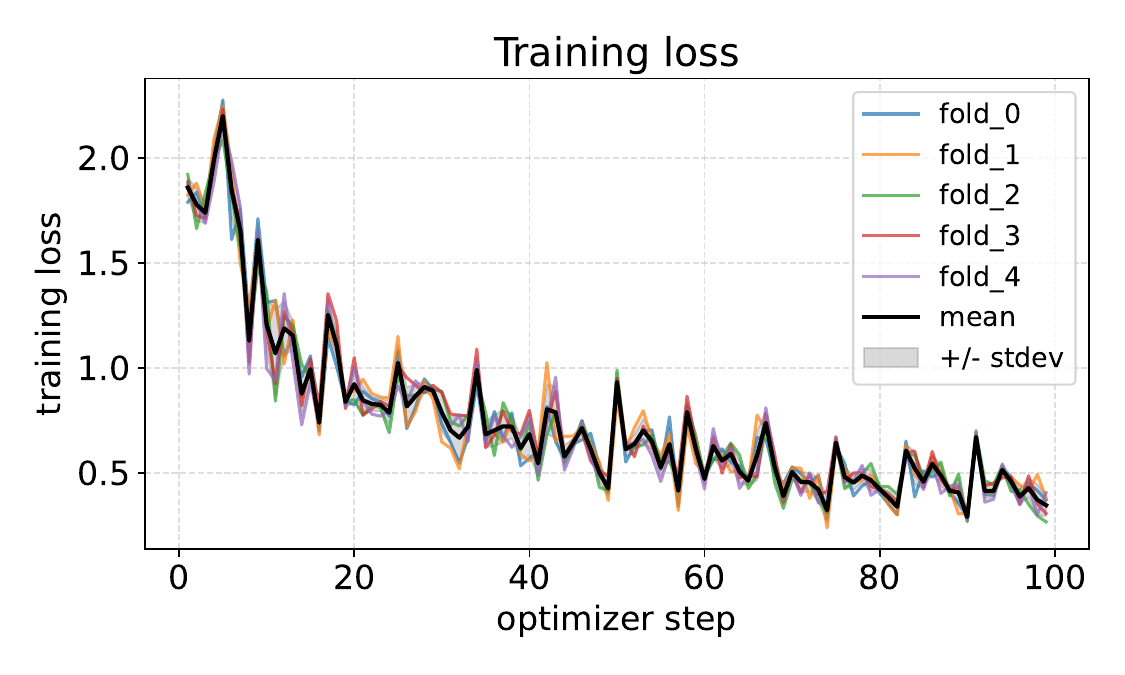}
  \caption{Fine-tuning loss curve on Gemma-3-4B-it.}
  \label{fig:finetuning_loss_gemma}
\end{figure}

\paragraph{Implementation Details.}
We fine-tuned \textit{Qwen2.5-1.5B-Instruct} and \textit{Gemma-3-4B-it} using QLoRA \cite{dettmers2023qloraefficientfinetuningquantized} with 4-bit NF4 quantization, double quantization, and fp16 compute precision. LoRA adapters (rank $16$, $\alpha=32$, dropout $0.05$) were applied to all attention and MLP projection layers. Training was performed for 3 epochs using an effective batch size of $16$, a cosine learning-rate schedule with peak learning rate $2\times10^{-4}$, and the \texttt{paged\_adamw\_8bit} optimizer. Experiments used 2048-token sequences, gradient checkpointing, and SDPA attention. All runs were conducted on a single NVIDIA Tesla T4 (16 GB) GPU. Figure~\ref{fig:finetuning_loss} and \ref{fig:finetuning_loss_gemma} shows the training loss curve for Qwen-2.5 and Gemma-3 respectively.

\paragraph{Evaluation.}
We compare the fine-tuned model against its corresponding base model. Results in Table~\ref{tab:finetuning_main} and Table~\ref{tab:finetuning_distortion_type} report the mean and standard deviation across all five cross-validation folds for the fine-tuned model, while the baseline results correspond to greedily decoded outputs from the original base model.

\section{Prompts Used}
\label{appx:prompt_details}
We have used standard prompt across all models and evaluation setups. For finetuned and open-source models, we followed prompt directions that were presented in their model card (TableGPT2-7B\footnote{\url{https://huggingface.co/tablegpt/TableGPT2-7B}}, TableLLM-7B\footnote{\url{https://huggingface.co/RUCKBReasoning/TableLLM-7b}}) to better reflect their capability. We used the prompt in Figure~\ref{fig:prompt_distortion_unaware} for Distortion-Unaware and prompt in Figure~\ref{fig:prompt_distortion_aware} for a Distortion-Aware setting. For the setting where we do not provide a code execution sandbox to the models, we use prompts in Figure~\ref{fig:prompt_distortion_unaware_no_sandbox} and \ref{fig:prompt_distortion_aware_no_sandbox} respectively.

\section{Accuracy Metric}
\label{appx:acc_metric}
We claim that exact match accuracy is a close proxy for succeful repair. This is given from the fact that cases where a model could recall the WikiTQ answer from pretraining were negligible.

For code-sandbox models, memorization would not help. A recalled script would not execute correctly on the distorted table because the distortions change the computation path. Although a model could have hard-coded the answer, we found no such cases after checking all logs.

For non-sandbox models such as Deepseek-R1-Distill Qwen-32B, we used an LLM-as-judge to inspect successful reasoning traces and check whether the model changed its computation or applied a repair under distortion when compared to its canonical execution. The judge flagged 93.88\% of 278 successful cases as explicit repairs. For the remaining 17, manual review showed that the model still applied a fix, such as associating a missing column name, without stating it explicitly. Thus, we found no correct answer produced without repair.

%Distortion Unaware prompt
\begin{figure*}[htbp]
\centering
\begin{tcolorbox}[
  width=0.95\linewidth,
  colback=gray!5,
  colframe=black,
  boxrule=0.4pt,
  arc=2pt
]
\small\ttfamily
You are an expert data analyst specializing in solving complex data analytics questions.
Your role is to analyze datasets and provide accurate answers to user queries.

\textbf{Your Environment:}
\begin{itemize}
  \item You have access to a code sandbox where you can execute Python scripts
  \item The relevant dataset (Excel or CSV file) has been uploaded to the sandbox
  \item You can write and run Python code to analyze the data and answer queries
\end{itemize}

\textbf{Your Responsibilities:}
\begin{enumerate}
  \item Carefully analyze the user's query to understand the required information
  \item Study the table structure and content to ensure accurate reasoning
  \item Write clear and efficient Python code to extract insights
  \item Execute the code in the sandbox environment
  \item If execution fails, debug and retry up to five times
  \item Provide a direct answer based on the analysis without suggesting approaches or partial solutions
\end{enumerate}
\end{tcolorbox}
\caption{Distortion-Unaware prompt given to models during evaluation.}
\label{fig:prompt_distortion_unaware}
\end{figure*}

% Distortion-UnAware Prompt no sandbox
\begin{figure*}[htbp]
\centering
\begin{tcolorbox}[
  width=0.95\linewidth,
  colback=gray!5,
  colframe=black,
  boxrule=0.4pt,
  arc=2pt
]
\small\ttfamily

You are an expert data analyst specializing in solving data analytics questions based on a tabular data. Your role is to analyze the data and provide accurate answers to user queries.

\textbf{Your Environment:}
\begin{itemize}
    \item You will be provided with the relevant data context required to answer the query.
\end{itemize}

\textbf{Your Responsibilities}:
\begin{enumerate}
    \item Carefully analyze the user's query to understand what information they need
    \item Study the table structure and content to better understand how to answer queries accurately
    \item Provide a direct answer based on your analysis. Do not suggest approaches or offer partial solutions
\end{enumerate}
\end{tcolorbox}
\caption{Distortion-UnAware prompt given to models during evaluation without the code execution sandbox.}
\label{fig:prompt_distortion_unaware_no_sandbox}
\end{figure*}

% Distortion-Aware Prompt
\begin{figure*}[htbp]
\centering
\begin{tcolorbox}[
  width=0.95\linewidth,
  colback=gray!5,
  colframe=black,
  boxrule=0.4pt,
  arc=2pt
]
\small\ttfamily
You are an expert data analyst specializing in solving complex data analytics questions.
Your role is to analyze datasets and provide accurate answers to user queries.

\textbf{Your Environment:}
\begin{itemize}
  \item You have access to a code sandbox where you can execute Python scripts
  \item The relevant dataset (Excel or CSV file) has been uploaded to the sandbox
  \item You can write and run Python code to analyze the data and answer queries
\end{itemize}

\textbf{Your Responsibilities:}
\begin{enumerate}
  \item Carefully analyze the user's query to understand the required information
  \item Study the table structure and content to ensure accurate reasoning
  \item Write clear and efficient Python code to extract insights from the dataset
  \item Execute the code in the sandbox environment
  \item If execution fails, debug and retry up to five times before concluding failure
  \item Provide a direct answer based on the analysis without suggesting approaches or partial solutions
  \item If the table exhibits structural or semantic inconsistencies (e.g., shifted rows or columns, semantic misalignment), correct these issues prior to analysis
\end{enumerate}
\end{tcolorbox}
\caption{Distortion-Aware prompt given to models during evaluation.}
\label{fig:prompt_distortion_aware}
\end{figure*}

% Distortion-Aware Prompt no sandbox
\begin{figure*}[htbp]
\centering
\begin{tcolorbox}[
  width=0.95\linewidth,
  colback=gray!5,
  colframe=black,
  boxrule=0.4pt,
  arc=2pt
]
\small\ttfamily

You are an expert data analyst specializing in solving data analytics questions based on a tabular data. Your role is to analyze the data and provide accurate answers to user queries.

\textbf{Your Environment:}
\begin{itemize}
    \item You will be provided with the relevant data context required to answer the query.
\end{itemize}

\textbf{Your Responsibilities}:
\begin{enumerate}
    \item Carefully analyze the user's query to understand what information they need
    \item Study the table structure and content to better understand how to answer queries accurately
    \item Provide a direct answer based on your analysis. Do not suggest approaches or offer partial solutions
    \item If you encounter scenarios where the table seems incorrect either structurally or semantically (e.g., shifted rows, shifted columns, semantic misalignment, etc.), correct these issues first before proceeding with your analysis.
\end{enumerate}
\end{tcolorbox}
\caption{Distortion-Aware prompt given to models during evaluation without the code execution sandbox.}
\label{fig:prompt_distortion_aware_no_sandbox}
\end{figure*}

%% file: custom.bib
@misc{bhandari2025exploringrobustnesslanguagemodels,
      title={Exploring the Robustness of Language Models for Tabular Question Answering via Attention Analysis}, 
      author={Kushal Raj Bhandari and Sixue Xing and Soham Dan and Jianxi Gao},
      year={2025},
      eprint={2406.12719},
      archivePrefix={arXiv},
      primaryClass={cs.CL},
      url={https://arxiv.org/abs/2406.12719}, 
}

@misc{zhang2025contentdifferentrepresentationscontrolled,
      title={Same Content, Different Representations: A Controlled Study for Table QA}, 
      author={Yue Zhang and Seiji Maekawa and Nikita Bhutani},
      year={2025},
      eprint={2509.22983},
      archivePrefix={arXiv},
      primaryClass={cs.CL},
      url={https://arxiv.org/abs/2509.22983}, 
}

@misc{su2024tablegpt2largemultimodalmodel,
      title={TableGPT2: A Large Multimodal Model with Tabular Data Integration}, 
      author={Aofeng Su and Aowen Wang and Chao Ye and Chen Zhou and Ga Zhang and Gang Chen and Guangcheng Zhu and Haobo Wang and Haokai Xu and Hao Chen and Haoze Li and Haoxuan Lan and Jiaming Tian and Jing Yuan and Junbo Zhao and Junlin Zhou and Kaizhe Shou and Liangyu Zha and Lin Long and Liyao Li and Pengzuo Wu and Qi Zhang and Qingyi Huang and Saisai Yang and Tao Zhang and Wentao Ye and Wufang Zhu and Xiaomeng Hu and Xijun Gu and Xinjie Sun and Xiang Li and Yuhang Yang and Zhiqing Xiao},
      year={2024},
      eprint={2411.02059},
      archivePrefix={arXiv},
      primaryClass={cs.LG},
      url={https://arxiv.org/abs/2411.02059}, 
}

@misc{pasupat2015compositionalsemanticparsingsemistructured,
      title={Compositional Semantic Parsing on Semi-Structured Tables}, 
      author={Panupong Pasupat and Percy Liang},
      year={2015},
      eprint={1508.00305},
      archivePrefix={arXiv},
      primaryClass={cs.CL},
      url={https://arxiv.org/abs/1508.00305}, 
}

@misc{zhang2025tablellmenablingtabulardata,
      title={TableLLM: Enabling Tabular Data Manipulation by LLMs in Real Office Usage Scenarios}, 
      author={Xiaokang Zhang and Sijia Luo and Bohan Zhang and Zeyao Ma and Jing Zhang and Yang Li and Guanlin Li and Zijun Yao and Kangli Xu and Jinchang Zhou and Daniel Zhang-Li and Jifan Yu and Shu Zhao and Juanzi Li and Jie Tang},
      year={2025},
      eprint={2403.19318},
      archivePrefix={arXiv},
      primaryClass={cs.CL},
      url={https://arxiv.org/abs/2403.19318}, 
}

@misc{sui2024tablemeetsllmlarge,
      title={Table Meets LLM: Can Large Language Models Understand Structured Table Data? A Benchmark and Empirical Study}, 
      author={Yuan Sui and Mengyu Zhou and Mingjie Zhou and Shi Han and Dongmei Zhang},
      year={2024},
      eprint={2305.13062},
      archivePrefix={arXiv},
      primaryClass={cs.CL},
      url={https://arxiv.org/abs/2305.13062}, 
}

@misc{wu2025realhitbenchcomprehensiverealistichierarchical,
      title={RealHiTBench: A Comprehensive Realistic Hierarchical Table Benchmark for Evaluating LLM-Based Table Analysis}, 
      author={Pengzuo Wu and Yuhang Yang and Guangcheng Zhu and Chao Ye and Hong Gu and Xu Lu and Ruixuan Xiao and Bowen Bao and Yijing He and Liangyu Zha and Wentao Ye and Junbo Zhao and Haobo Wang},
      year={2025},
      eprint={2506.13405},
      archivePrefix={arXiv},
      primaryClass={cs.CL},
      url={https://arxiv.org/abs/2506.13405}, 
}

@misc{qwen2.5-VL,
    title = {Qwen2.5-VL},
    url = {https://qwenlm.github.io/blog/qwen2.5-vl/},
    author = {Qwen Team},
    month = {January},
    year = {2025}
}

@misc{deepseekai2025deepseekr1incentivizingreasoningcapability,
      title={DeepSeek-R1: Incentivizing Reasoning Capability in LLMs via Reinforcement Learning}, 
      author={DeepSeek-AI},
      year={2025},
      eprint={2501.12948},
      archivePrefix={arXiv},
      primaryClass={cs.CL},
      url={https://arxiv.org/abs/2501.12948}, 
}

@misc{openai2026gpt54,
  author = {{OpenAI}},
  title = {Introducing GPT-5.4},
  year = {2026},
  howpublished = {\url{https://openai.com/index/introducing-gpt-5-4/}},
  note = {Accessed: 2026-05-22}
}

@misc{anthropic2026opus46,
  author = {{Anthropic}},
  title = {Claude Opus 4.6},
  year = {2026},
  howpublished = {\url{https://www.anthropic.com/news/claude-opus-4-6}},
  note = {Accessed: 2026-05-22}
}

@inproceedings{chen-etal-2020-hybridqa,
    title = "{H}ybrid{QA}: A Dataset of Multi-Hop Question Answering over Tabular and Textual Data",
    author = "Chen, Wenhu  and
      Zha, Hanwen  and
      Chen, Zhiyu  and
      Xiong, Wenhan  and
      Wang, Hong  and
      Wang, William Yang",
    editor = "Cohn, Trevor  and
      He, Yulan  and
      Liu, Yang",
    booktitle = "Findings of the Association for Computational Linguistics: EMNLP 2020",
    month = nov,
    year = "2020",
    address = "Online",
    publisher = "Association for Computational Linguistics",
    url = "https://aclanthology.org/2020.findings-emnlp.91/",
    doi = "10.18653/v1/2020.findings-emnlp.91",
    pages = "1026--1036"
}

@inproceedings{yavuz-etal-2018-takes,
    title = "What It Takes to Achieve 100{\%} Condition Accuracy on {W}iki{SQL}",
    author = "Yavuz, Semih  and
      Gur, Izzeddin  and
      Su, Yu  and
      Yan, Xifeng",
    editor = "Riloff, Ellen  and
      Chiang, David  and
      Hockenmaier, Julia  and
      Tsujii, Jun{'}ichi",
    booktitle = "Proceedings of the 2018 Conference on Empirical Methods in Natural Language Processing",
    month = oct # "-" # nov,
    year = "2018",
    address = "Brussels, Belgium",
    publisher = "Association for Computational Linguistics",
    url = "https://aclanthology.org/D18-1197/",
    doi = "10.18653/v1/D18-1197",
    pages = "1702--1711"
}

@inproceedings{Herzig_2020,
   title={TaPas: Weakly Supervised Table Parsing via Pre-training},
   url={http://dx.doi.org/10.18653/v1/2020.acl-main.398},
   DOI={10.18653/v1/2020.acl-main.398},
   booktitle={Proceedings of the 58th Annual Meeting of the Association for Computational Linguistics},
   publisher={Association for Computational Linguistics},
   author={Herzig, Jonathan and Nowak, Pawel Krzysztof and Müller, Thomas and Piccinno, Francesco and Eisenschlos, Julian},
   year={2020} }

@misc{liu2022tapextablepretraininglearning,
      title={TAPEX: Table Pre-training via Learning a Neural SQL Executor}, 
      author={Qian Liu and Bei Chen and Jiaqi Guo and Morteza Ziyadi and Zeqi Lin and Weizhu Chen and Jian-Guang Lou},
      year={2022},
      eprint={2107.07653},
      archivePrefix={arXiv},
      primaryClass={cs.CL},
      url={https://arxiv.org/abs/2107.07653}, 
}

@misc{jiang2022omnitabpretrainingnaturalsynthetic,
      title={OmniTab: Pretraining with Natural and Synthetic Data for Few-shot Table-based Question Answering}, 
      author={Zhengbao Jiang and Yi Mao and Pengcheng He and Graham Neubig and Weizhu Chen},
      year={2022},
      eprint={2207.03637},
      archivePrefix={arXiv},
      primaryClass={cs.CL},
      url={https://arxiv.org/abs/2207.03637}, 
}

@misc{zheng2024multimodaltableunderstanding,
      title={Multimodal Table Understanding}, 
      author={Mingyu Zheng and Xinwei Feng and Qingyi Si and Qiaoqiao She and Zheng Lin and Wenbin Jiang and Weiping Wang},
      year={2024},
      eprint={2406.08100},
      archivePrefix={arXiv},
      primaryClass={cs.CL},
      url={https://arxiv.org/abs/2406.08100}, 
}

@misc{yang2025doestablesourcematter,
      title={Does Table Source Matter? Benchmarking and Improving Multimodal Scientific Table Understanding and Reasoning}, 
      author={Bohao Yang and Yingji Zhang and Dong Liu and André Freitas and Chenghua Lin},
      year={2025},
      eprint={2501.13042},
      archivePrefix={arXiv},
      primaryClass={cs.CL},
      url={https://arxiv.org/abs/2501.13042}, 
}

@inproceedings{zhao-etal-2023-robut,
    title = "{R}obu{T}: A Systematic Study of Table {QA} Robustness Against Human-Annotated Adversarial Perturbations",
    author = "Zhao, Yilun  and
      Zhao, Chen  and
      Nan, Linyong  and
      Qi, Zhenting  and
      Zhang, Wenlin  and
      Tang, Xiangru  and
      Mi, Boyu  and
      Radev, Dragomir",
    editor = "Rogers, Anna  and
      Boyd-Graber, Jordan  and
      Okazaki, Naoaki",
    booktitle = "Proceedings of the 61st Annual Meeting of the Association for Computational Linguistics (Volume 1: Long Papers)",
    month = jul,
    year = "2023",
    address = "Toronto, Canada",
    publisher = "Association for Computational Linguistics",
    url = "https://aclanthology.org/2023.acl-long.334/",
    doi = "10.18653/v1/2023.acl-long.334",
    pages = "6064--6081"
}

@misc{ye2025tableqameetsnoisedual,
      title={When TableQA Meets Noise: A Dual Denoising Framework for Complex Questions and Large-scale Tables}, 
      author={Shenghao Ye and Yu Guo and Dong Jin and Yikai Shen and Yunpeng Hou and Shuangwu Chen and Jian Yang and Xiaofeng Jiang},
      year={2025},
      eprint={2509.17680},
      archivePrefix={arXiv},
      primaryClass={cs.CL},
      url={https://arxiv.org/abs/2509.17680}, 
}

@misc{zhou2025tablequestionansweringera,
      title={Table Question Answering in the Era of Large Language Models: A Comprehensive Survey of Tasks, Methods, and Evaluation}, 
      author={Wei Zhou and Bolei Ma and Annemarie Friedrich and Mohsen Mesgar},
      year={2025},
      eprint={2510.09671},
      archivePrefix={arXiv},
      primaryClass={cs.CL},
      url={https://arxiv.org/abs/2510.09671}, 
}

@misc{zhu2025tableevalrealworldbenchmarkcomplex,
      title={TableEval: A Real-World Benchmark for Complex, Multilingual, and Multi-Structured Table Question Answering}, 
      author={Junnan Zhu and Jingyi Wang and Bohan Yu and Xiaoyu Wu and Junbo Li and Lei Wang and Nan Xu},
      year={2025},
      eprint={2506.03949},
      archivePrefix={arXiv},
      primaryClass={cs.CL},
      url={https://arxiv.org/abs/2506.03949}, 
}

@inproceedings{balakrishnan2015applying,
  title={Applying WebTables in Practice.},
  author={Balakrishnan, Sreeram and Halevy, Alon Y and Harb, Boulos and Lee, Hongrae and Madhavan, Jayant and Rostamizadeh, Afshin and Shen, Warren and Wilder, Kenneth and Wu, Fei and Yu, Cong},
  booktitle={CIDR},
  year={2015}
}

@INPROCEEDINGS{5277546,
  author={Oro, Ermelinda and Ruffolo, Massimo},
  booktitle={2009 10th International Conference on Document Analysis and Recognition}, 
  title={PDF-TREX: An Approach for Recognizing and Extracting Tables from PDF Documents}, 
  year={2009},
  volume={},
  number={},
  pages={906-910},
  doi={10.1109/ICDAR.2009.12}}

@misc{dettmers2023qloraefficientfinetuningquantized,
      title={QLoRA: Efficient Finetuning of Quantized LLMs}, 
      author={Tim Dettmers and Artidoro Pagnoni and Ari Holtzman and Luke Zettlemoyer},
      year={2023},
      eprint={2305.14314},
      archivePrefix={arXiv},
      primaryClass={cs.LG},
      url={https://arxiv.org/abs/2305.14314}, 
}

@article{humanannotation,
author = {Feinstein, Alvan and Cicchetti, Domenic},
year = {1990},
month = {02},
pages = {543-9},
title = {Feinstein AR, Cicchetti DVHigh agreement but low kappa: I. The problems of two paradoxes. J Clin Epidemiol 43:543-549},
volume = {43},
journal = {Journal of clinical epidemiology},
doi = {10.1016/0895-4356(90)90158-L}
}
